 \documentclass[letterpaper, 10 pt, conference]{ieeeconf}  

\IEEEoverridecommandlockouts 

\usepackage{graphics} 
\usepackage{amsmath} 
\usepackage{graphicx}
\graphicspath{ {./images/} } 

\usepackage[
backend=biber,
style=numeric,
sorting=nyvt,
]{biblatex}

\usepackage{tabularx}
\usepackage{caption}
\usepackage{subcaption}
\usepackage{comment}

\DeclareCaptionSubType * [alph]{table}
\begin{document}

\title{\LARGE \bf
Towards Efficient 3D Object Detection in Bird's-Eye-View Space for Autonomous Driving: A Convolutional-Only Approach
}

\author{Yuxin Li$^{1,2}$ Qiang Han $^{2}$ Mengying Yu$^{2}$  Yuxin Jiang$^{1}$ Chai Kiat Yeo$^{1}$\\ 
Yiheng Li$^{2}$ Zihang Huang $^{2}$ Nini Liu $^{2}$ Hsuanhan Chen $^{2}$ Xiaojun Wu$^{2}$ 
}

\maketitle

\thispagestyle{empty}
\pagestyle{empty}

\begin{abstract}
3D object detection in Bird's-Eye-View (BEV) space has recently emerged as a prevalent approach in the field of autonomous driving. Despite the demonstrated improvements in accuracy and velocity estimation compared to perspective view methods, the deployment of BEV-based techniques in real-world autonomous vehicles remains challenging. This is primarily due to their reliance on vision-transformer (ViT) based architectures, which introduce quadratic complexity with respect to the input resolution. To address this issue, we propose an efficient BEV-based 3D detection framework called BEVENet, which leverages a convolutional-only architectural design to circumvent the limitations of ViT models while maintaining the effectiveness of BEV-based methods. Our experiments show that BEVENet is 3$\times$ faster than contemporary state-of-the-art (SOTA) approaches on the NuScenes challenge, achieving a mean average precision (mAP) of 0.456 and a nuScenes detection score (NDS) of 0.555 on the NuScenes validation dataset, with an inference speed of 47.6 frames per second. To the best of our knowledge, this study stands as the first to achieve such significant efficiency improvements for BEV-based methods, highlighting their enhanced feasibility for real-world autonomous driving applications.
\end{abstract}

\section{Introduction}
3D object detection in Bird's-Eye-View (BEV) space has gained considerable traction within the autonomous driving research community. As an alternative to LiDAR-based methods, generating pseudo-LiDAR points using surrounding cameras has emerged as a cost-effective and promising solution in the domain of autonomous driving. Consequently, numerous approaches \cite{li2022bevformer, huang2021bevdet, huang2022bevdet4d, li2022bevdepth, philion2020lift, cape2023xiong, solofusion2022park, tigbev2022huang} have been proposed to incorporate perception tasks into the BEV space.


However, existing methods are typically computationally demanding and heavily reliant on large-scale datasets. While these conditions can be met within laboratory settings, they present considerable obstacles to implementation in real-world, in-vehicle environments. The vision transformer (ViT) module \cite{dosovitskiy2020vit}, is the primary component responsible for substantial GPU memory consumption and matrix operations. Although the ViT architecture is widely utilized in BEV-based methods due to its capacity to capture global semantic information, it necessitates training on extensive datasets and requires a significantly longer training time than convolutional neural networks (CNNs) to facilitate the model's understanding of the positional relationships between pixels. Despite the increased training costs, ViT offers only marginal improvements on various vision benchmarks compared to CNN-based models.
\begin{figure}[t]
\centering
\includegraphics[width=0.45\textwidth]{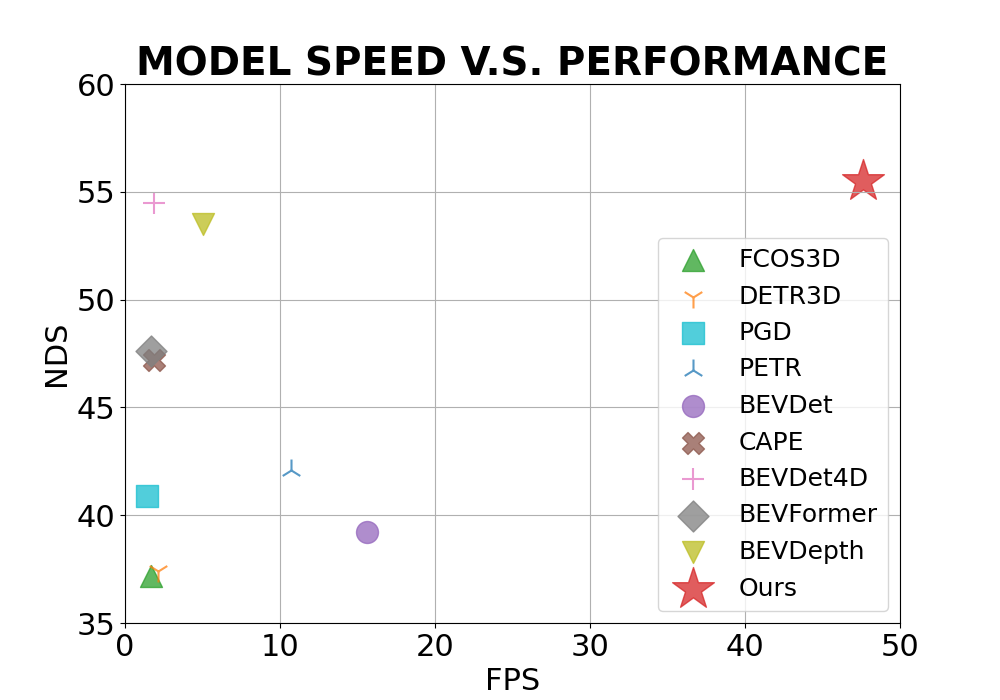}
\caption{The comparison of inference speed over different state-of-the-art methods}
\label{fig:speed}
\end{figure}

Another noteworthy limitation of ViT models is their quadratic complexity with respect to input dimensions, specifically the resolution of the input image. While these models are undoubtedly powerful, their deployment on embedded devices is hindered by constrained computing resources. In addition, large input resolution is certainly favoured by ViT models to increase the performance of 3D detection. However, the majority of objects in autonomous driving scenes are relatively small and consequently their detection remains a persistent challenge for ViT models.

\begin{figure*}[t]
\centering
\includegraphics[width=0.8\textwidth]{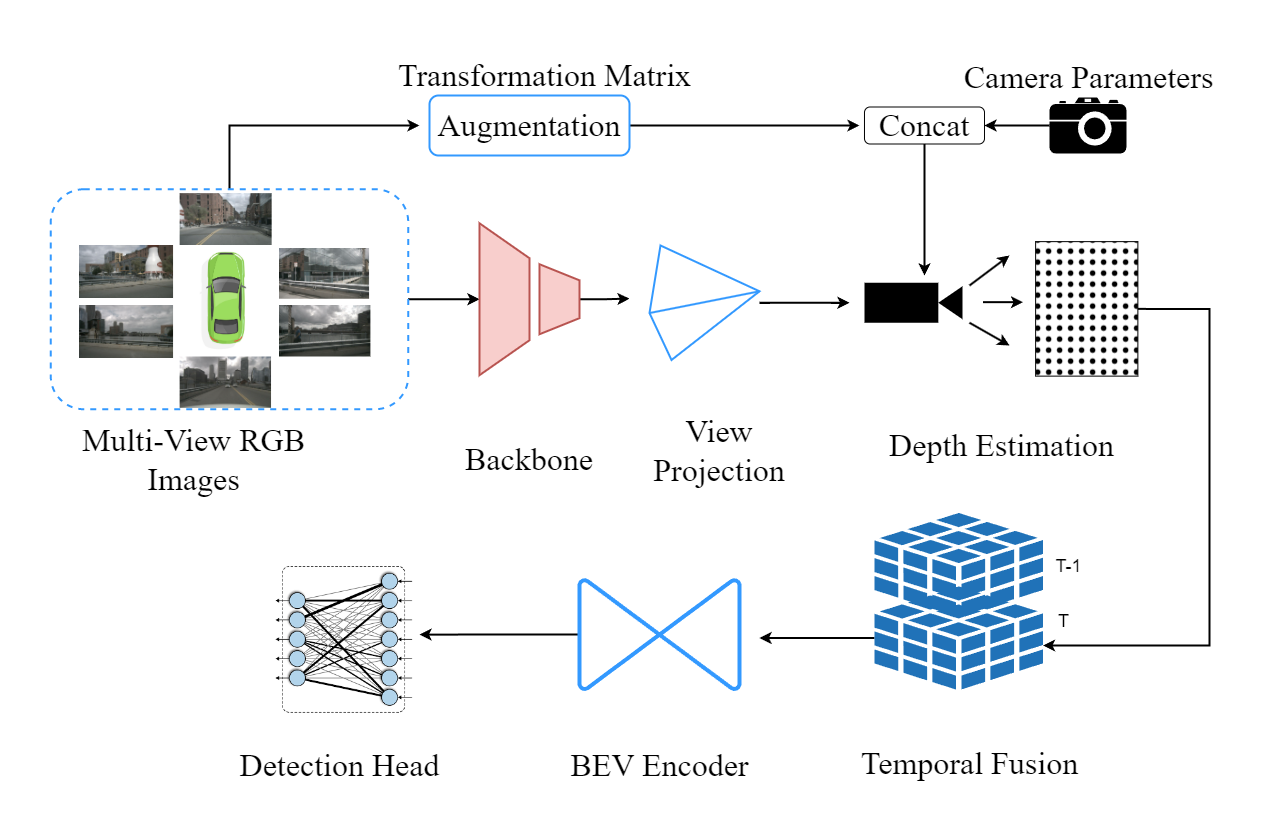}
\caption{Overall Architecture of BEVENet. BEVENet consists of six major modules: Backbone, View Projector, Depth Estimator, Temporal Fuser, BEV Encoder and Detection Head. During the inference stage, only multi-view camera input is needed in the pipeline, whereas during training, LiDAR points are included as a rich source of supervision signals for the depth estimation module. Refer to Figure \ref{fig:depth} for more details. }
\label{fig:bevnet}
\end{figure*}
Based on the aforementioned analysis, we propose to address these limitations by investigating alternative approaches, such as purely CNN-based modules. In this work, our primary objective is to design an efficient 3D detection framework that employs the BEV paradigm under constrained hardware conditions. To this end, we systematically analyze six fundamental components in the 3D detection pipeline: backbone, view projection, depth estimation, temporal fusion, BEV feature encoding and detection head. Model complexity and benchmarking metrics are all taken into account in the analysis given that they are essential metrics for the real-world deployment of the neural network models.

We specifically propose BEVENet, an abbreviation for BEV-Efficient-Neural-Network, as a resource-efficient model designed for the real-world deployment of BEV-based methods. By adopting a convolutional-only design, we aim to accelerate the model's inference speed while maintaining comparable performance to state-of-the-art (SOTA) methods. As illustrated in Figure \ref{fig:speed}, we demonstrate that a purely CNN-based implementation of the BEV architecture serves as a robust alternative to transformer-based models. With the best-reported performance at mAP = 0.456 and NDS = 0.555, our model achieves an inference speed of 47.6 frames per second, which is three times faster and nearly ten times smaller in GFlops than contemporary SOTA methods on the NuScenes challenge. 

\section{Related work}
\subsection{Backbone Models}
Backbone models are the cornerstones of visual perception tasks. The inception of AlexNet \cite{alex2012alexnet} ignited a rapid progression of advancements in various vision tasks, with VGGNet \cite{olga2015vgg} and ResNet \cite{He2015} representing the first two significant milestones in backbone models for vision applications. VGGNet initially demonstrated the feasibility of enhancing vision task performance by increasing the depth of neural networks, while ResNet further expanded the depth of backbone models to 152 layers through the innovative use of skip connections. The Deformable Convolutional Network (DCN) \cite{dai2017dcn} revealed that fixed filter sizes are not obligatory for learning visual features.

In the realm of efficient inference, EfficientNet \cite{tan2019efficient} initially underscored the importance of deployment-friendly design, while RepVGG \cite{ding2021repvgg} recently established a burgeoning paradigm in inference-oriented models. Moreover, ElanNet \cite{wang2022yolov7}, a vital embodiment of structural reparameterization, has showcased the unparalleled advantages of CNN models in deployment scenarios.

Beyond the field of computer vision, natural language processing (NLP) tasks have also reaped the benefits of advancements in backbone models. Transformers and their variants have dominated a considerable number of NLP tasks. Inspired by the widespread application of transformers in NLP, Dosovitskiy et al. proposed the Vision Transformer (ViT) \cite{dosovitskiy2020vit} as a universal backbone model for computer vision. Drawing insights from dilated convolutional layers, Swin Transformer \cite{liu2021Swin} represents another remarkable attempt to apply ViT to vision tasks.

\begin{figure}[t]
\centering
\includegraphics[width=0.4\textwidth]{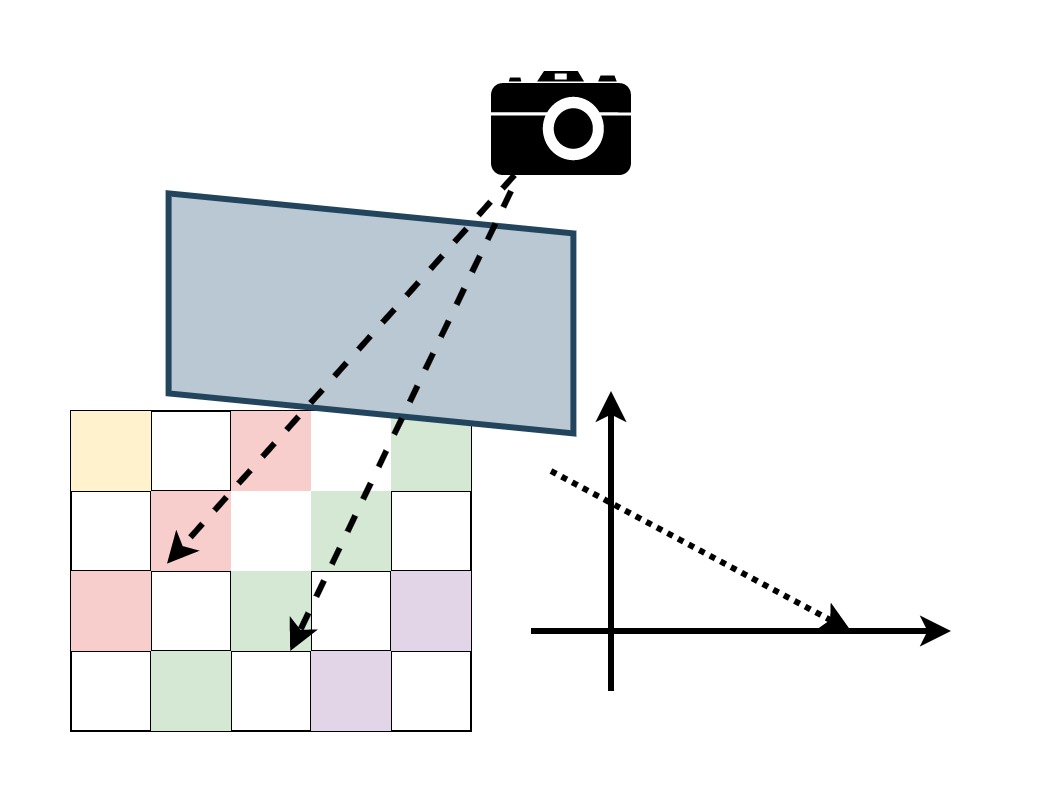}
\caption{Illustration of View Projection. Camera images from the 2D domain are lifted to the 3D space along the light ray; projection is made in both the horizontal and vertical directions.}
\label{fig:lss}
\end{figure}

\subsection{3D Detection Methods}
3D object detection, a fundamental task within the perception modules of autonomous driving systems, has experienced substantial advancements since the introduction of the BEV paradigm. Leveraging the precise semantic information provided by surrounding cameras, temporal input from previous frames, and depth information supervision from LiDAR inputs, camera-only 3D detection methods  \cite{li2022bevdepth, li2022bevformer, huang2021bevdet, huang2022bevdet4d, cape2023xiong, solofusion2022park, tigbev2022huang} are rapidly approaching the performance of their LiDAR-based counterparts.

DETR3D \cite{detr3d} pioneered the use of a multi-view image input paradigm to enhance the performance of 3D detection. Incorporating temporal information, BEVFormer \cite{li2022bevformer} achieved a 10-point increase in mAP compared to DETR3D. BEVDet4D \cite{huang2021bevdet}, which extended Lift-Splat-Shoot (LSS) \cite{philion2020lift} with an explicit BEV-Encoder, achieved similar results without requiring additional input from CAN bus information. BEVDepth \cite{li2022bevdepth}, which incorporated depth supervision from LiDAR input, further validated the effectiveness of perception within the BEV space. Lift-Splat-Shoot \cite{philion2020lift}, a classic yet efficient method for projecting images from a 2D-view to a 3D-view, initiated the utilization of surrounding camera input to enhance 3D perception performance. Each image is individually "lifted" into a frustum of features for each camera, "splat" all frustums into a rasterized bird's-eye view grid, and finally "shoot" different trajectories onto the cost map. With the emergence of BEV-based methods, it is anticipated that pure-vision-based approaches will soon match the performance of LiDAR-based methods.

\section{Methodology}
\subsection{Design Philosophy}
Our objective is to design an efficient model tailored for deployment on limited hardware resources while maintaining the precision of BEV-based methods. We adopt a reduction-based methodology, iteratively reducing the complexity of each module. Specifically, we first disassemble the SOTA methods on the NuScenes challenge leaderboard through theoretical decomposition and complexity analysis by GFlop. Subsequently, we iteratively combine alternatives for each module, prioritizing speed as the benchmark for design choices. Finally, we try to improve the performance of the final 3D detection task by combining optimal model-tuning tactics from well-established baselines.

\subsection{Network Structure}
As depicted in Figure \ref{fig:bevnet}, the BEVENet architecture  comprises six modules: a shared backbone model ElanNet with NuImage pretraining; a view projection module LSS with lookup table; a fully-convolutional depth estimation module with data augmentation; a temporal module with 2-second historical information; a BEV feature encoder with residual blocks; and lastly, a simplified detection head with Circular NMS.

\begin{figure}[t]
\centering
\includegraphics[width=0.4\textwidth]{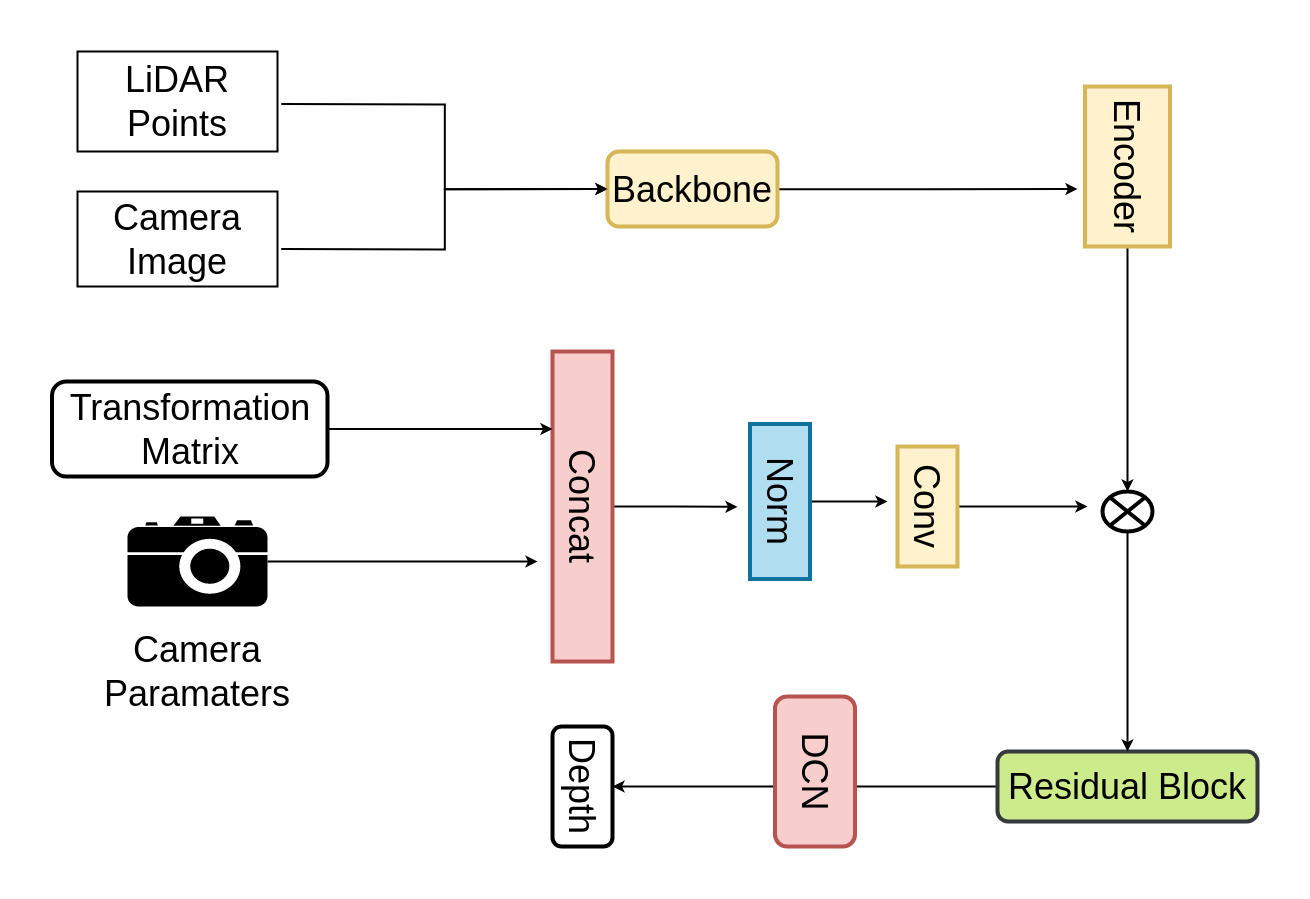}
\caption{Illustration of the Depth Module. We adopt the same design as BEVDepth \cite{li2022bevdepth} in depth estimation module, but add the augmentation matrix and extrinsic parameters together with the intrinsic parameters as input to the depth estimation network. The MLP layer is also being replaced by a convolutional network.}
\label{fig:depth}
\end{figure}

\subsection{Backbone Model}
Our backbone model forms the cornerstone of the 3D detection task, leveraging input from six surrounding view cameras to extract essential semantic features for the subsequent tasks. With the primary objective of mitigating the complexity challenges posed by the Vision Transformer (ViT) model, we embark on a comparative study using a carefully curated selection of four models.

In our quest to contrast the intricacies between ViT models and their convolutional counterparts, we strategically choose two representative models from each category. These include the naive Vision-Transformer (ViT) \cite{dosovitskiy2020vit}, Swin Transformer (SwinT) \cite{liu2021Swin}, the Resnet \cite{He2015} and ElanNet \cite{wang2022yolov7}. Through this comparative study, we aim to unearth the best-performing model while adhering to our goals of improved simplicity and enhanced performance. Apart from the effort to compare the backbones, we simultaneously investigate the possible techniques to improve the model performance; we tried to mitigate the performance deterioration caused by the statistical offset of datasets from different sources. Specifically, pretraining on NuImage \cite{nuscenes2019} has been adopted.

\begin{figure*}[t]
\begin{minipage}{0.45\textwidth}
    \centering
    \begin{subfigure}{\textwidth}
        \includegraphics[scale=0.85,width=0.85\textwidth]{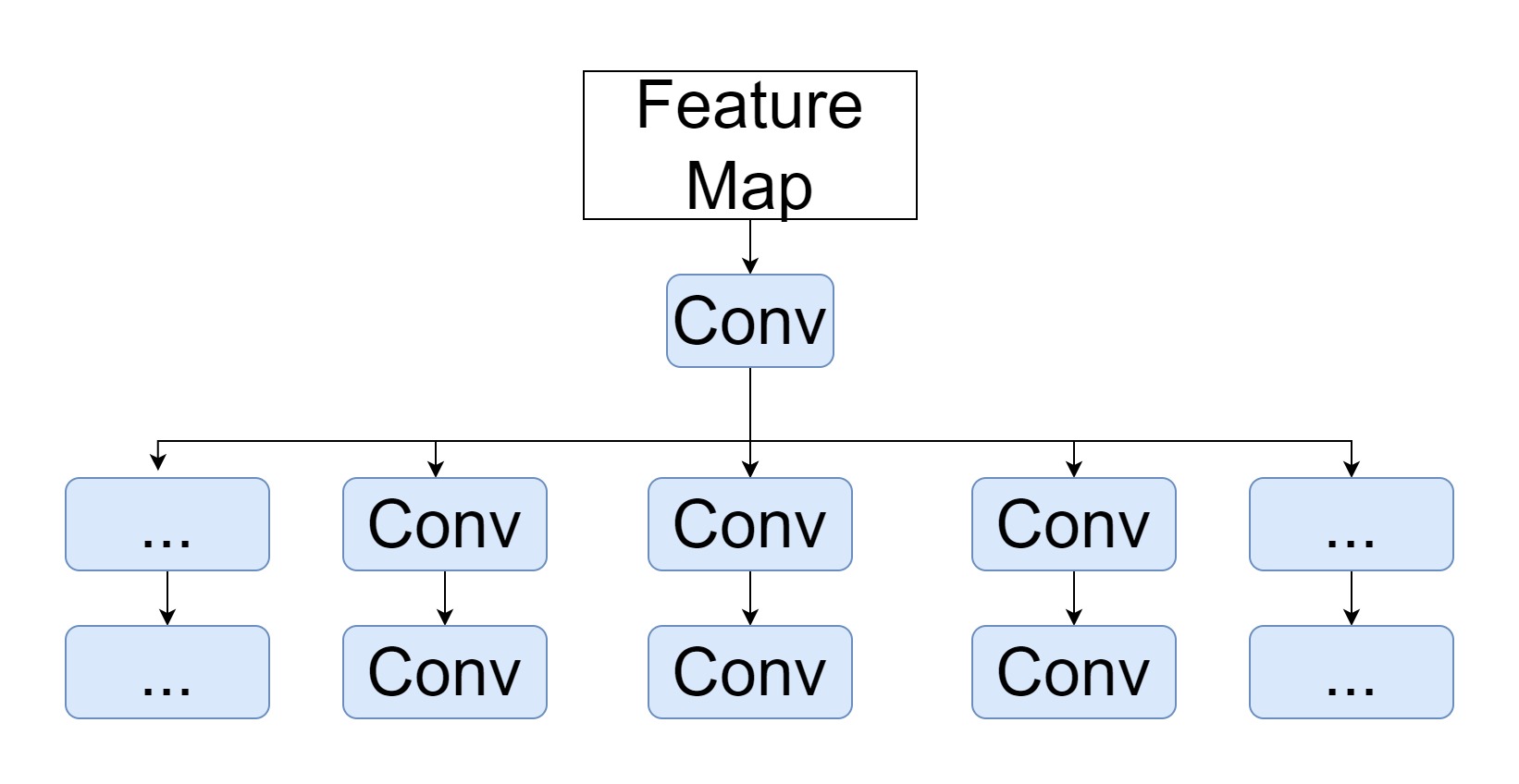}
        \caption{Original Head}
        \label{fig:orihead}
    \end{subfigure}
    \begin{subfigure}{\textwidth}
        \includegraphics[scale=0.85,width=0.85\textwidth]{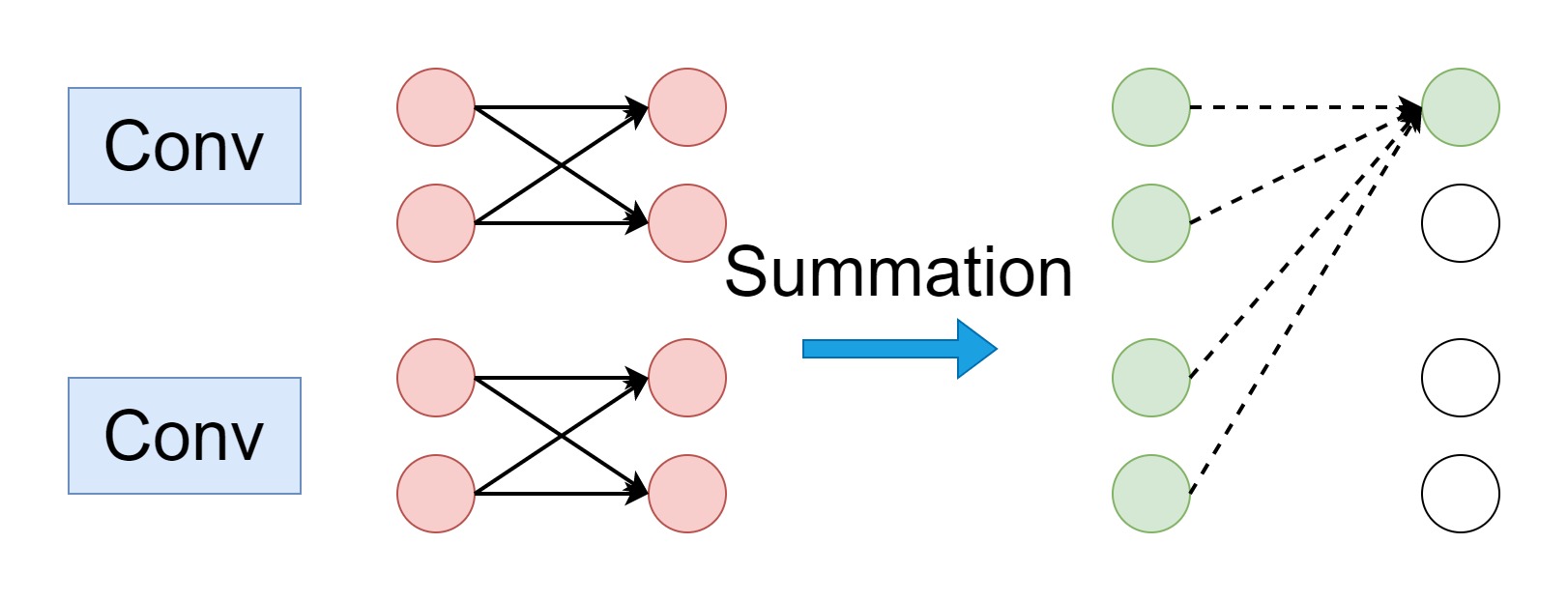}
        \caption{Merging Multi-Branch Parallel Conv Layers}
        \label{fig:mergeconv}
    \end{subfigure}
\end{minipage}\hfill
\begin{minipage}{0.45\textwidth}
    \centering
    \begin{subfigure}{\textwidth}
        \includegraphics[scale=0.85,width=0.85\textwidth]{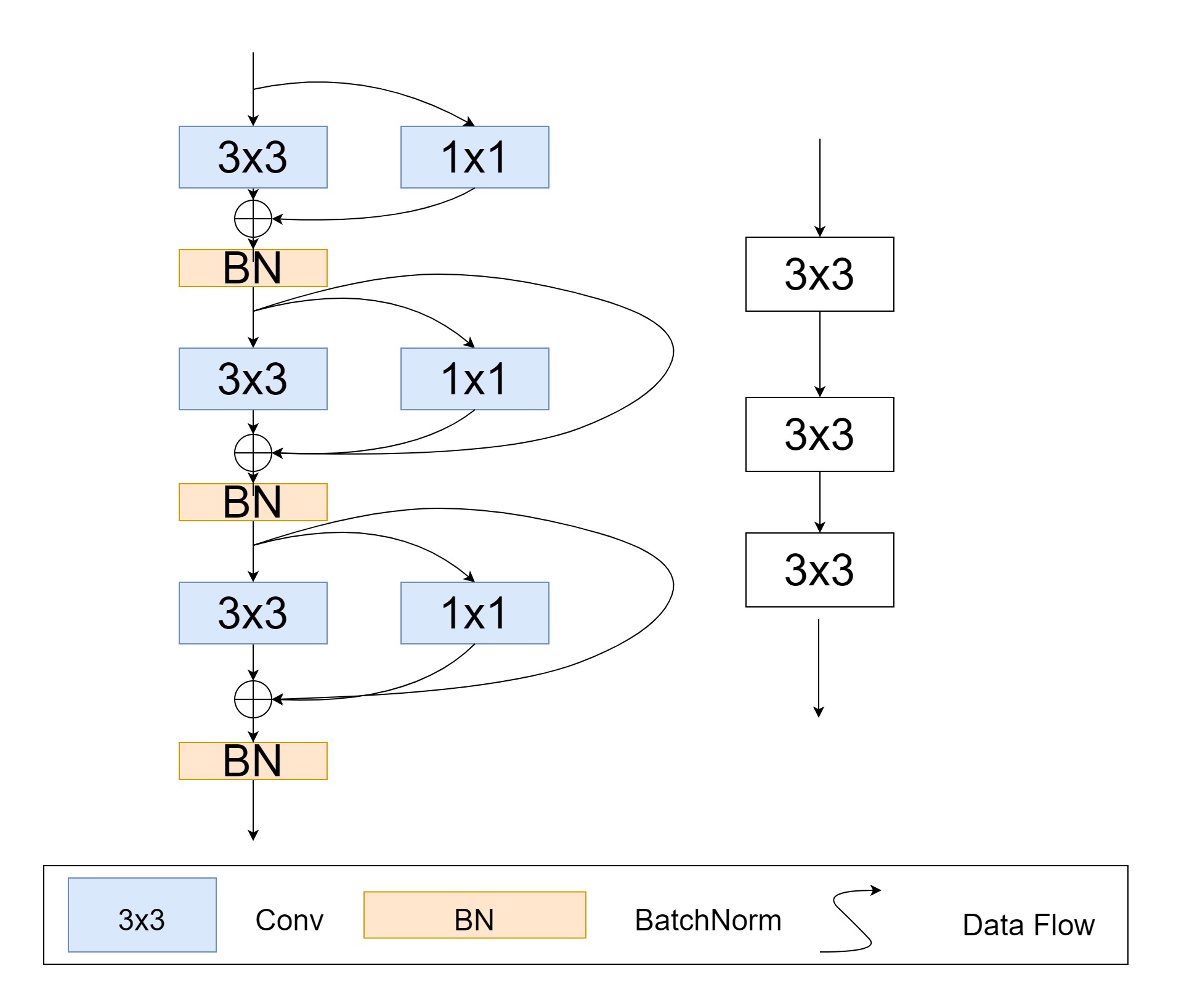}
        \caption{Merging Conv-BN Layers}
        \label{fig:convbn}
    \end{subfigure}
\end{minipage}
\caption{Illustration of Detection Head Simplification by Re-Parameterization. Compared to the original detection head, we combine the output nodes mathematically by their values, which will generate identical results but with fewer multiplication operations. }
\label{fig:comparehead}
\end{figure*}

\subsection{View Projection}
Following Lift-Splat-Shoot \cite{philion2020lift} and BEVDet \cite{huang2021bevdet}, our feature projection module predicts depth probability for each pixel, calculating ground truth depth based on geometric similarity (Figure \ref{fig:lss}). Pixel $p$'s coordinates $(u,v)$ are transformed into feature map vector $f(u,v)$, representing depth probability at position $f(u,v)$ with vector $alpha \times f(u,v)$. Through light ray projection and the camera's intrinsic and extrinsic parameters, image domain coordinates are converted to spatial domain. Final BEV feature computation happens via voxelization on a flat surface. The fixed 2D-to-3D projection matrix involves no learnable parameters. To accelerate inference, we compute this matrix during data processing and load it via a look-up table during training and testing.

\subsection{Depth Prediction}
A depth estimation module is introduced to compensate the depth accuracy noise caused by view projection. This module overlays the frustum cloud point depth with its own depth predictions, averaging the two via empirically-determined weights. The module ingests both LiDAR points and multi-view images; the former serves as ground truth, while the latter undergoes augmentation to strengthen prediction robustness (Figure \ref{fig:depth}). The amalgamation of image features, camera parameters, and the image augmentation transformation matrix are inputted into an encoding layer. The depth estimation module utilizes both the intrinsic and extrinsic camera parameters to enhance depth prediction.

\subsection{Temporal Fusion and BEV Encoder}
Our temporal fusion module, designed to augment 3D detection accuracy, leverages the model's ability to harness latent temporal information. In obscured or occluded scenarios, it can infer hidden objects' locations based on their past positions. This module, simply designed, processes accumulated feature maps from prior frames through a convolutional encoder, utilizing features from the preceding two-second span to better interpret obscured object motion and positioning.

Simultaneously, the BEV-Encoder module serves as an intermediate layer connecting the pseudo-LiDAR cloud and the final detection head. Two residual blocks are employed to transform the sparse LiDAR points into a dense matrix of feature points. Each grid in the BEV space is generated via voxelization with a predefined resolution.

\subsection{Detection Head}
Built upon the BEV feature, the detection head is based on CenterPoint \cite{yin2021center}. We set the prediction target to include the position, scale, orientation and speed of the objects in the autonomous driving scenes. We adopt the same setup as CenterPoint during the training stage for a fair comparison against other algorithms. Formally we set the loss to be
\begin{equation}
L_{\text{total}} = \frac{\alpha L_{\text{det}} + \beta L_{\text{cls}} + \gamma L_{\text{2D}}}{N}
\label{eq:loss}
\end{equation}
where $L_{\text{det}}$ is Smooth L1 Loss, $L_{\text{cls}}$ is Gaussian Focal Loss, $L_{\text{2D}}$ is also Smooth L1 Loss but is applied on the 2D input as a penalty to let the model improve its localization ability.

During the inference stage, as per RepVGG \cite{ding2021repvgg}, we re-parameterize all the multi-branch convolutional layers and batch-norm layers into cascaded plain convolutional networks. As depicted in Figure \ref{fig:orihead}, the detection head comprises several parallel convolutional neural networks. This structure can be simplified by merging the convolutional and batch-norm layers. As shown in Figure \ref{fig:convbn}, a ResNet-like architecture is equivalent to a plain convolutional neural network without skip-connection or 1x1 convolution. The identity module can be directly added to the output feature map without any special operation. Meanwhile, the batch-norm layer can be combined with the convolutional layer by mathematically summing the mean value and standard variance of the batch input. 

\begin{table*}[t]
\caption{Comparison against SOTA methods on the NuScenes dataset. Performance of the SOTA methods at resolutions other than 704 x 256 is lifted from the original papers due to the absence of the model weights in the code releases.}
\label{table:sota}
\begin{center}
\begin{tabular}{c|c c c|c|c| c c c c c}
\hline
Model & Image Size & GFlops $\downarrow$ & FPS $\uparrow$ & mAP $\uparrow$ & NDS $\uparrow$ & mATE $\downarrow$ & mASE $\downarrow$ & mAOE $\downarrow$ & mAVE $\downarrow$ & mAAE $\downarrow$ \\
\hline 
FCOS3D \cite{wang2021fcos3d}        & 1600x900 & 2008.2 & 1.7   & 29.5 & 37.2 & 0.806 & 0.268 & 0.511     & 1.315 & \bf0.170 \\
DETR3D \cite{detr3d}                & 1600x900 & 1016.8 & 2.1   & 30.3 & 37.4 & 0.860 & 0.278 & 0.437     & 0.967 & 0.235 \\
PGD \cite{wang2021pgd}              & 1600x900 & 2223.0 & 1.4   & 33.5 & 40.9 & 0.732 & 0.263 & 0.423     & 1.285 & 0.172 \\
BEVFormer \cite{li2022bevformer}    & 1600x900 & 1303.5 & 1.7   & 41.6 & 47.6 & 0.673 & 0.274 & 0.372     & 0.394 & 0.198 \\
CAPE                                & 1600x900 & - & -          & 43.9 & 47.9 & 0.683 & 0.267 & 0.427     & 0.814 & 0.197 \\
\hline 
BEVDet                              & 1600x640 & 2962.6 & 1.9   & 39.3 & 47.2 & 0.608 & 0.259 & 0.366     & 0.822 & 0.191 \\
BEVDet4D                            & 1600x640 & 2989.2 & 1.9   & 42.1 & 54.5 & 0.579 & 0.258 & \bf0.329  & 0.301 & 0.191 \\
\hline 
PETR \cite{liu2022petr}             & 1408x512 & - & 3.4        & 35.7 & 42.1 & 0.710 & 0.270 & 0.490     & 0.885 & 0.224 \\
BEVDepth                            & 1408x512 & - & 5.0        & 41.2 & 53.5 & 0.565 & 0.266 & 0.358     & 0.331 & 0.190 \\
BEVDepth+TiGBEV                     & 1408x512 & - & -          & 44.0 & 54.4 & 0.570 & 0.267 & 0.392     & 0.331 & 0.201 \\
\hline 
FCOS3D \cite{wang2021fcos3d}        & 704x256 & 223.1  & 17.2   & 23.3 & 30.1 & 0.911 & 0.281 & 0.731     & 1.259 & 0.219 \\
DETR3D \cite{detr3d}                & 704x256 & 195.5  & 19.7   & 23.9 & 31.0 & 0.972 & 0.292 & 0.625     & 1.073 & 0.304 \\
PGD \cite{wang2021pgd}              & 704x256 & 220.1  & 13.6   & 26.5 & 34.9 & 0.713 & \bf0.239 & 0.471     & 1.291 & 0.337 \\
PETR                                & 704x256 & 228.4  & 23.1   & 28.2 & 34.9 & 0.806 & 0.283 & 0.700     & 0.978 & 0.289 \\
BEVDet \cite{huang2021bevdet}       & 704x256 & 215.3  & 33.1   & 31.2 & 39.2 & 0.691 & 0.272 & 0.523     & 0.909 & 0.247 \\
CAPE                                & 704x256 & 207.1  & 13.3   & 31.8 & 44.2 & 0.760 & 0.277 & 0.560     & 0.386 & 0.182 \\
BEVFormer \cite{li2022bevformer}    & 704x256 & 217.2  & 19.9   & 32.8 & 39.5 & 0.661 & 0.259 & 0.357     & 1.593 & 0.197 \\
BEVDet4D \cite{huang2022bevdet4d}   & 704x256 & 222.0  & 15.5   & 33.8 & 47.6 & 0.672 & 0.274 & 0.460     & 0.337 & 0.185 \\
BEVDepth+TiGBEV                     & 704x256 & 233.7  & 14.7   & 35.6 & 47.7 & 0.648 & 0.273 & 0.517     & 0.364 & 0.210 \\
BEVDepth                            & 704x256 & 212.0  & 18.6   & 35.7 & 48.1 & 0.609 & 0.262 & 0.511     & 0.408 & 0.202 \\
SOLOFusion                          & 704x256 & 223.2  & 11.1   & 42.7 & 53.4 & 0.567 & 0.274 & 0.411     & \bf0.252 & 0.188 \\
BEVENet(Ours)                       & 704x256 & \bf161.42 & \bf47.6 & \bf45.6 & \bf55.5 & \bf0.549 & 0.278 & 0.438 & 0.270 & 0.196 \\
\hline
\end{tabular}
\end{center}
\end{table*}

Formally, let \(x_1 = w \times x_0 + b\) and \(x_2 = \frac{x_1 - mean}{std}\), then 
\begin{align}
x_2 &= \frac{w \times x_0 + b  - mean}{std} \\ 
&= (\frac{w}{std}) \times x_0 + (\frac{b-mean}{std}) \\
&= w' \times x_0 + b'
\end{align}
where $w'$ is the new weight combining the convolutional layer and batchnorm layer, and $b'$ is the new bias.

As illustrated in Figure \ref{fig:mergeconv}, the multi-branch convolutional module can be simplified by merging the weights and biases of the neighbouring cells. Assuming that performance degradation at a single cell prior to the merge operation is bounded by post-quantization error $E$, the maximum number of mergeable cells $n$ is determined by the pre-quantization or full-precision quantization error $e$ divided by the post-quantization error, $n=E/e$. The number of neighbouring cells to merge is empirically set to two in our experiments.

\section{EXPERIMENTS}
\subsection{Experiment Settings}
\textbf{Dataset and Evaluation Metrics}
Our model, BEVENet, is evaluated using the NuScenes benchmark dataset \cite{nuscenes2019}, encompassing 1,000 driving scenes captured via six cameras and a LiDAR sensor. This dataset annotates ten classes within a 51.2-meter ground plane for the 3D detection task. Performance evaluation leverages both official NuScenes metrics, namely, mean Average Precision (mAP), Average Translation Error (ATE), Average Scale Error (ASE), Average Orientation Error (AOE), Average Velocity Error (AVE), Average Attribute Error (AAE), and NuScenes Detection Scores (NDS), as well as efficiency-oriented metrics, namely, Frame Per Second (FPS) and GFlops. The former measures performance on NVIDIA A100 GPU with pre-processing and post-processing time excluded, while the latter uses the MMDetection3D \cite{mmdet3d2020} toolkit.

\textbf{Training Parameters}
Models are trained to utilize the Adam optimizer with a learning rate of 2e-4 and weight decay set at 1e-2. The batch size is configured as 4 on each of the 8 A100 GPU cards. We train for 24 epochs per experimental round and save the best-performing model for evaluation.

\textbf{Data Processing}
Our data processing adopts an approach similar to BEVDet \cite{huang2021bevdet}, tailored for the specific needs of the NuScenes dataset which has an original resolution of $1600 \times 900$. We rescale this to $704 \times 256$ during preprocessing. Key transformations include random flipping, scaling, cropping, rotation, and a copy-paste mechanism to address any skewness in object distribution. These augmentation operations are mathematically converted into transformation matrices, combined with camera parameters that are flattened for dimension consistency. Class-Balanced-Grouping-and-Sampling (CBGS) \cite{zhu2019class}, in concert with the copy-paste mechanism, is applied during training, following the methodology of CenterPoint \cite{yin2021center}. In testing phase, images are scaled only but not cropped to align with the model's input dimensions.

\subsection{Performance Benchmark}
We select 11 SOTA methods on NuScenes leaderboard as our baselines: BEVFormer \cite{li2022bevformer}, BEVDet \cite{huang2021bevdet}, BEVDet4D \cite{huang2022bevdet4d}, BEVDepth \cite{li2022bevdepth}, PETR \cite{liu2022petr}, PGD \cite{wang2021pgd}, FCOS3D \cite{wang2021fcos3d}, DETR3D \cite{detr3d}, CAPE \cite{cape2023xiong}, SoloFusion \cite{solofusion2022park} and TiGBEV \cite{tigbev2022huang}. 

From Table \ref{table:sota}, we can see that BEVENet achieves significant improvement in various performance measures when compared to the SOTA methods. With an image size of 704x256, BEVENet outperforms all the other models in computational efficiency at the lowest GFlops of 161.42. This reflects the resource efficiency of BEVENet, making it particularly suitable for deployment in hardware-constrained environments. In terms of FPS, BEVENet also excels with a frame rate of 47.6. As for other core performance metrics, BEVENet achieves an mAP of 45.6 and NDS of 55.5, again, the highest among all the models.

In terms of the five error metrics, BEVENet exhibits the lowest mATE of 0.549, testifying to its superior performance in translation estimation. Although BEVENet's mASE, mAOE and mAAE are not the absolute lowest, they remain competitive against the other models. For mAVE, BEVENet demonstrates a commendable performance with a relatively low error of 0.270, only bested by SOLOFusion.

In summary, BEVENet stands out amongst the current SOTA models, showcasing exceptional performance in terms of computational efficiency and core detection metrics while maintaining competitive error rates. This evaluation validates BEVENet's potential as a viable and efficient solution for 3D object detection in real-world autonomous driving scenarios.

\section{Ablation Study}
In the above sections, we have shown our model's best performance capabilities, juxtaposed against other SOTA models. In this section, we present the analytical journey in identifying the modules which land up in our final design via a complexity analysis. Given the varied outcomes from the different module configurations, we present the rationale behind our decision. 

To kickstart the analysis, we first proposed six SOTA baseline configurations for each of the six major modules of BEVENet shown in Figure 2. These baseline configurations are ResNet50, LSS, original BEVDepth, BEV encoder with a visual transformer model, a temporal fusion window set to eight seconds and a detection head leveraging the CenterHead with Scale-NMS. These baseline configurations are highlighted in bold in Table \ref{table:ablation}. We shall call this initial BEVENet used for complexity analysis as BEVENet-Baseline to distinguish it from our final proposed BEVENet. The additions to these baselines are marked with '+'. These details aim to provide a solid foundation for our upcoming complexity analysis and design rationales.

\begin{figure}[t]
\centering
\includegraphics[width=0.4\textwidth]{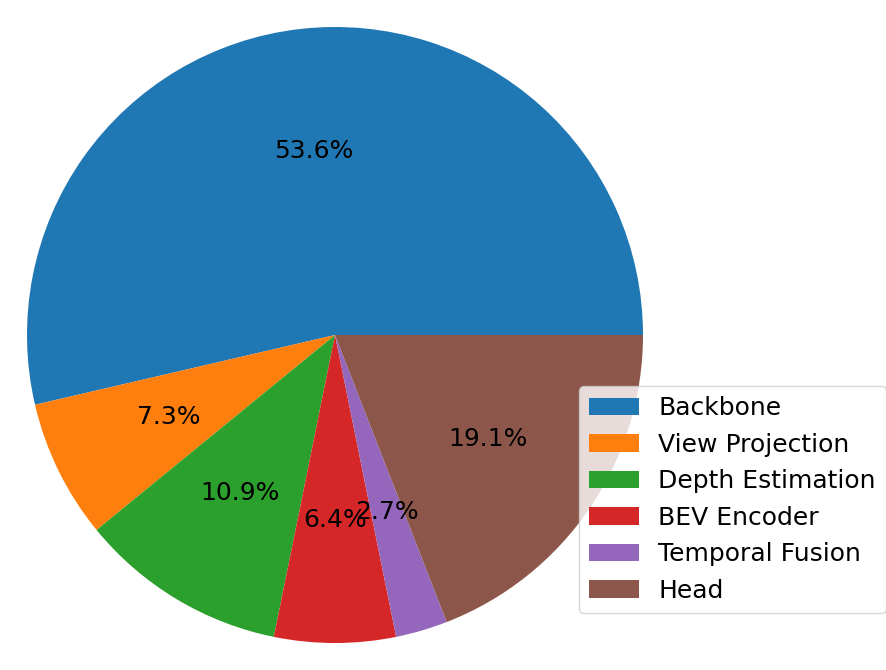}
\caption{Percentage of GFlops of Each Module. GFlops are measured using the mmdet3d toolkit. The sequence of each module is presented in an anti-clockwise manner.}
\label{fig:gflops}
\end{figure}

\subsection{Complexity Analysis}
Aiming to design an efficient 3D detection framework for resource-constrained hardware, we conducted a complexity analysis from two perspectives to pinpoint the most computationally intensive segments.

\textbf{Modular Complexity Analysis}
Complexity analysis on the modules of BEVENet-Baseline (Figure \ref{fig:gflops}) shows the backbone module, detection head, and depth estimation module consume over 80\% of GFlops during inference, necessitating strategies for GFlops reduction. Input resolution, a critical complexity determinant, was examined using benchmarks from FCOS3D, DETR3D, PGD, BEVFormer, and CAPE (Table \ref{table:sota}). Reducing input resolution from $1600\times900$ to $704\times256$ significantly lowered complexity, albeit with initial performance degradation, suggesting the need for other performance-boosting techniques.

\textbf{Complexity Analysis on Views}
Considering the variable relevance of each of the six surrounding-view cameras on autonomous vehicles, we analyzed the effect of view omission on GFlops reduction. Our findings (Table \ref{table:maskingview}) indicate that the front and back views impact the model's mAP and NDS more than the side views, with NDS showing more sensitivity to view masking. This suggests that masking back views could efficiently reduce model complexity with minimal performance loss.

\begin{table}[t]
\caption{Masking Camera Views vs Performance}
\label{table:maskingview}
\begin{center}
\begin{tabular}{c|c|c}
\hline
Mask & NDS & mAP\\
\hline
Baseline & 35.4 & 25.22 \\
\hline
Front & 24.46 & 19.23 \\
Front Right & 23.60 & 24.42 \\
Front Left & 23.65 & 24.33 \\
Back Right & 34.36 & 23.60 \\
Back Left & 34.40 & 23.94 \\
Back & 22.01 & 17.14 \\
\hline
Front Left \& Right & 23.30 & 18.92 \\
Back Left \& Right & 31.72 & 19.13 \\
\hline
Back Three Cams &  17.84 & 10.89 \\
Side Four Cams & 16.89 & 10.98 \\
\hline
\end{tabular}\\
\end{center}
\end{table}

\subsection{Backbone Module}
In analyzing different backbone models, we scrutinized ViT \cite{dosovitskiy2020vit}, SwinTransformer \cite{liu2021Swin}, ResNet \cite{He2015}, and ELanNet \cite{wang2022yolov7}, each with a similar parameter count for impartial comparison. Starting with ResNet50 as our baseline, with FPS at 27.4, mAP at 39.3 and NDS at 45.9 as per Table \ref{subtab:backboneablation}, we found ViT underperforming in FPS at 17.9, Swin-T on par with ResNet50, and ELanNet superior in FPS at 30.2. Augmenting ELanNet with NuImage pretraining enhanced performance notably, increasing FPS to 30.3 while improving mAP and NDS to 42.0 and 50.1, respectively. Therefore, ELanNet, designed for inference efficiency, outpaced ViT, Swin-Transformer, and ResNet50 in this task.

\begin{table*}
\caption{Design Alternatives vs Performance. The baseline for each stage is shown in bold while the additions to the baseline are marked with '+'. Note that when varying the designs of the current stage, the configurations which yield the best results in all its preceding stages are being adopted.}
\label{table:ablation}
    \begin{subtable}{0.48\textwidth}
    \caption{Backbone Module}
    \label{subtab:backboneablation}
    \centering
        \begin{tabular}{c|c|c|c}
            \hline
            Backbone & FPS & mAP & NDS \\
            \hline
            \textbf{Resnet50} & 27.4 & 39.3 & 45.9 \\
            \hline
            Naive ViT & 17.9 & 30.9 & 38.7 \\            
            \hline
            Swin-T  & 27.1 & 38.5 & 44.4\\
            \hline
            ElanNet & 30.2 & 39.6 & 45.8 \\
            \hline
            + NuImage Pretraining & 30.3 & 42.0 & 50.1 \\
            \hline
        \end{tabular} 
    \end{subtable}
    \begin{subtable}{0.48\textwidth}
    \caption{View Projection Module.}
    \label{subtab:projectionablation}
    \centering
        \begin{tabular}{c|c|c|c}
            \hline
            View Projection & FPS & mAP & NDS \\
            \hline
            \textbf{LSS} & 30.3 & 42.0 & 50.1 \\
            \hline
            Transformer  & 29.9 & 42.1 & 50.0\\
            \hline
            MLP & 31.5 & 39.8 & 46.2\\
            \hline
            LSS + Lookup Table & 34.9 & 42.1 & 50.1 \\
            \hline
        \end{tabular} 
    \end{subtable}
    \medskip
    
    \begin{subtable}{0.48\textwidth}
    \caption{Depth Estimation Module}
    \label{subtab:depthablation}
    \centering
        \begin{tabular}{c|c|c|c}
            \hline
            Depth Estimation & FPS & mAP & NDS \\
            \hline
            \textbf{BEVDepth} & 34.9 & 42.1 & 50.1 \\
            \hline
            + Residual Encoder  & 35.8 & 43.9 & 52.7\\
            \hline
            + Image Aug Params & 35.7 & 44.5 & 53.3 \\
            \hline
        \end{tabular} 
    \end{subtable}
    \begin{subtable}{0.48\textwidth}
    \caption{Temporal Fusion Module}
    \label{subtab:temporalablation}
    \centering
        \begin{tabular}{c|c|c|c}
            \hline
            Temporal Fusion & FPS & mAP & NDS \\
            \hline
            \textbf{+ 8s} & 35.7 & 44.5 & 53.3 \\
            \hline
            + 6s & 35.8 & 44.5 & 53.3 \\
            \hline
            + 4s & 36.1 & 44.5 & 53.2 \\
            \hline
            + 2s & 36.3 & 44.4 & 53.2 \\
            \hline
        \end{tabular} 
    \end{subtable}
    \medskip

    \begin{subtable}{0.48\textwidth}
    \caption{BEV Encoder Module}
    \label{subtab:bevencoderablation}
    \centering
        \begin{tabular}{c|c|c|c}
            \hline
            BEV Encoder & FPS & mAP & NDS \\
            \hline
            \textbf{Transformer} & 36.3 & 44.4 & 53.2 \\
            \hline
            MLP  & 35.7 & 34.7 & 43.3\\
            \hline
            Residual Blocks & 38.8 & 44.2 & 52.5 \\
            \hline
        \end{tabular} 
    \end{subtable}
    \begin{subtable}{0.48\textwidth}
    \caption{Detection Head}
    \label{subtab:detheadablation}
    \centering
        \begin{tabular}{c|c|c|c}
            \hline
            Detection Head & FPS & mAP & NDS \\
            \hline
            \textbf{CenterHead(Scale-NMS)} & 38.8 & 44.2 & 52.5 \\
            \hline
            Circular NMS & 43.1 & 43.9 & 51.7 \\
            \hline
            + 2D Loss  & 43.0 & 45.5 & 55.5\\
            \hline
            + Re-parameterization & 47.6 & 45.6 & 55.5 \\
            \hline
        \end{tabular} 
    \end{subtable}
\end{table*}

\subsection{View Projection and Depth Estimation Modules}
Our view projection module's capability, which is pivotal for 2D-to-3D transformation, was examined across various configurations, with ElanNet featuring NuImage pretraining as our backbone. Despite minimal FPS, mAP, and NDS disparities between LSS, Transformer, and MLP methods, as seen in Table \ref{subtab:projectionablation}, LSS, with pre-calculated image to point cloud conversion matrix, provided a notable boost in FPS to 34.9, verifying its effectiveness for view projection.

Simultaneously, the depth estimation module, vital for post-projection depth optimization, was examined. Initially mirroring BEVDepth structure \cite{li2022bevdepth}, Figure \ref{fig:gflops} revealed its substantial complexity contribution of 10.9\%. Consequently, the MLP layer was replaced with a 2-layer residual block, a minimalist redesign elevating FPS by nearly one point and mAP by two points (Table \ref{subtab:depthablation}). Moreover, we further enhanced it by integrating the image augmentation matrix. Despite the seemingly marginal impact, our experimental results confirmed the efficacy of this strategy in bolstering model performance.

\subsection{Temporal Fusion and BEV Encoder Modules}
The temporal fusion module, crucial for efficacious inference in high-occlusion environments and velocity estimation refinement, was evaluated over a range of temporal window lengths. As exhibited in Table \ref{subtab:temporalablation}, shortening the interval from "8s" to "2s" modestly elevated FPS by 0.6 without appreciable detriment to mAP or NDS scores.

Simultaneously, the BEV encoder module, which acts as a liaison between temporally amalgamated features and the detection head, unexpectedly enhanced FPS by two frames, despite its modest complexity as seen in Figure \ref{fig:gflops}. Starting with Transformer as the baseline, we experimented with MLP and Residual Blocks replacements. Table \ref{subtab:bevencoderablation} attests to the superior performance of Residual Blocks, elevating FPS to 38.8 and thus endorsing their integration within our design.

\subsection{Detection Head}
As a culminating step, we proceeded to refine the detection head. CenterHead with Scale-NMS, our original choice for the detection head, accounted for nearly 20\% of the model's complexity. Typically, a crucial post-processing step in 3D object detection heads is the application of Non-Maximum Suppression (NMS) to eliminate redundant bounding boxes. In our baseline configuration, we discovered that Scale-NMS could be supplanted by Circular NMS, incurring only a minor accuracy trade-off. As demonstrated in Table \ref{subtab:detheadablation}, this replacement resulted in an FPS increase of 4.3 frames, with negligible impact on accuracy. To counter the minor performance decrease induced by Circular NMS, we introduced 2D supervision on the object bounding box in the image domain, a strategy that successfully elevated the mAP to 45.5 and NDS to 55.5. Our final enhancement entailed the re-parameterization of the head, maintaining mathematical output equivalence but considerably simplifying the topology. Ultimately, this yielded our model's peak performance, with FPS at 47.6, mAP at 45.6, and NDS at 55.5.

\section{CONCLUSIONS}
In this paper, we introduced BEVENet, a novel and efficient framework specifically designed for 3D object detection under the computational constraints of real-world autonomous driving systems. BEVENet is architectured exclusively around convolutional layers within neural networks, a design choice that distinguishes it from other models in this realm. To the best of our knowledge, this is the first work that integrates the limitations of hardware resources on real-world autonomous driving vehicles within the BEV paradigm. For future work, we will explore the roles of various image inputs in a multi-view setup and to study the significance of Region-of-Interest within BEV to further improve the performance and efficiency of our model.



\clearpage

\printbibliography
\end{document}